\pdfoutput=1

\documentclass[11pt]{article}

\usepackage[final]{acl}

\usepackage{times}
\usepackage{latexsym}

\usepackage[T1]{fontenc}

\usepackage[utf8]{inputenc}

\usepackage{microtype}

\usepackage{inconsolata}

\usepackage{graphicx}

\usepackage{amsmath}
\usepackage{amsthm}
\usepackage{booktabs}
\usepackage{algorithm}
\usepackage{algorithmic}
\usepackage{amssymb}
\usepackage{multirow}
\usepackage{float}
\usepackage{stfloats}
\usepackage{placeins} 
\usepackage{listings} 
\usepackage{xcolor} 
\usepackage{wrapfig}
\usepackage{xurl}
\usepackage{hyperref}

\usepackage{fancyhdr}
\title{CoAT: Chain-of-Associated-Thoughts Framework for Enhancing Large Language Models Reasoning}

\author{
 \textbf{Jianfeng Pan}, 
 \textbf{Senyou Deng},
 \textbf{Shaomang Huang}
\\
\\
 360 Security Technology Inc.
 Beijing, China
\\
 \small{
   \textbf{Emails:} \href{mailto:panjianfeng@360.cn}{panjianfeng@360.cn}, \href{mailto:dengsenyou@360.cn}{dengsenyou@360.cn},
   \href{mailto:huangshaomang@360.cn}{huangshaomang@360.cn}
 }
}

\begin{document}
\maketitle
\thispagestyle{fancy}

\begin{abstract}
Research on LLM technologies is rapidly emerging, with most of them employ a `fast thinking' approach to inference. Most LLMs generate the final result based solely on a single query and LLM’s reasoning capabilities. However, with the advent of OpenAI-o1, `slow thinking' techniques have garnered increasing attention because its process is closer to the human thought process.
Inspired by the human ability to constantly associate and replenish knowledge during thinking, we developed the novel $\bf{C}$hain-$\bf{o}$f-$\bf{A}$ssociated-$\bf{T}$houghts (CoAT) framework,
which introduces an innovative synergy between the Monte Carlo Tree Search (MCTS) algorithm and a dynamic mechanism for integrating new key information, termed `associative memory'. By combining the structured exploration capabilities of MCTS with the adaptive learning capacity of associative memory, CoAT significantly expands the LLM search space, enabling our framework to explore diverse reasoning pathways and dynamically update its knowledge base in real-time. This allows the framework to not only revisit and refine earlier inferences but also adaptively incorporate evolving information, ensuring that the final output is both accurate and comprehensive. 
We validate CoAT’s effectiveness across a variety of generative and reasoning tasks. Quantitative experiments show that CoAT achieves over 10\% performance improvement on open-source multi-hop reasoning datasets (HotpotQA, MuSiQue) and more than 15\% gain on our proprietary CRB dataset.
\end{abstract}

\begin{figure}[t]
    \centering
    \includegraphics[width=0.85\linewidth]{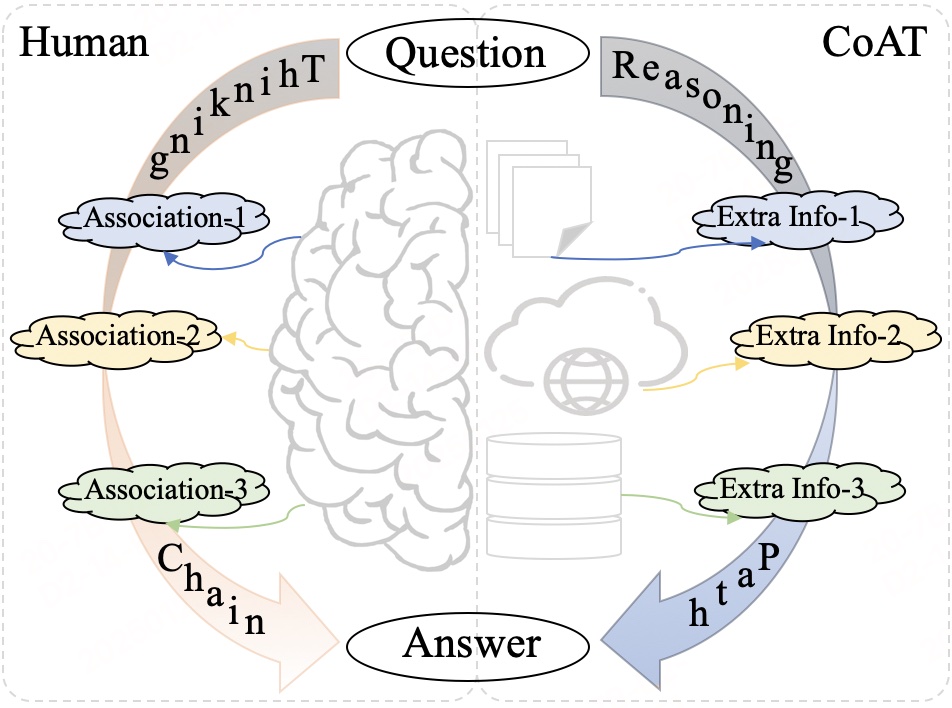}
    \caption{Left: Human thinking chain; Right: Associated thoughts path. This figure illustrates how our CoAT framework is inspired to continually supplement extra information during reasoning by simulating human associative mechanisms.}
    \label{fig:simula}
\end{figure}

\section{Introduction}
Large Language Models (LLMs) have rapidly become a cornerstone in natural language processing, powering applications ranging from conversational agents to complex decision-making systems. Central to their operation is the process of inference, where LLMs generate contents 
by an auto-regressive learning algorithm in the pre-training stage. Most LLMs, such as GPT~\cite{achiam2023gpt}, LLaMA~\cite{dubey2024llama}, and Qwen~\cite{yang2024qwen2}, employ a `fast thinking' approach to inference
which relies heavily on the pre-trained reasoning capabilities of LLM models. These approaches process a single query to produce the final result. Although effective for many tasks, they often struggle with problems that require nuanced, iterative reasoning, or adaptation to new information.

Recent advances~\cite{li2022competition,brown2024large,wu2025inference} have begun to explore alternatives to `fast thinking', introducing `slow thinking' methodologies~\cite{jiang2024technical,min2024imitate,gan2025rethinking} that align more closely with human thinking processes. This idea emphasizes deliberate, iterative reasoning, and the integration of historical contents or external knowledge during inference. OpenAI-o1~\cite{jaech2024openai}, a notable project, has sparked significant interest in this domain, showcasing the potential of `slow thinking' frameworks to improve reasoning capabilities.
Some studies~\cite{zhang2023planning,liu2023don,choi2023kcts,chen2024alphamath,tian2024toward,zhang2024rest,wan2024alphazero} have employed MCTS-inspired methods to enhance the multi-step reasoning capabilities of LLMs.
However, the above mentioned methods merely subdivide the reasoning process into smaller steps and involve rethinking what has already been generated. Throughout the process, reliance is still placed on the initial input information and the logical reasoning abilities of the LLM itself.

Inspired by the human ability to constantly associate and replenish knowledge during thinking, we propose the Chain-of-Associated-Thoughts (CoAT) framework.
To our knowledge, associative memory mechanisms were first applied to simulate human thoughts in LLM processes. The associative memory mechanism empowers CoAT to dynamically incorporate new key information during inference, mimicking the human ability to associate and update knowledge iteratively. 
Furthermore, we optimize the routing strategy in the MCTS algorithm to ensure that each addition of associative memory will provide additional key information for subsequent content generation.
This synergy between structured search and adaptive learning enables CoAT to expand its reasoning scope while maintaining contextual coherence, overcoming limitations of conventional LLMs. 

The effectiveness of our framework is validated through extensive experiments.
The results demonstrate that our framework significantly outperforms traditional models in terms of accuracy, coherence, and diversity.
In summary, the main contributions of our work are as follows:
\begin{itemize}
    \item We propose the CoAT framework to enhance LLM reasoning. Our framework expands the LLM reasoning search space for a better solution using the optimized MCTS algorithm.
    \item We endow the LLM reasoning process with human-like associative and adaptive self-refinement capabilities to effectively address complex reasoning tasks.
    \item We optimize the routing strategy in CoAT to identify the best generation trajectory. 
    The qualitative and quantitative experimental results demonstrate its superior performance compared to other methods.
\end{itemize}


\begin{figure*}[t]
    \centering
    \includegraphics[width=0.85\textwidth]{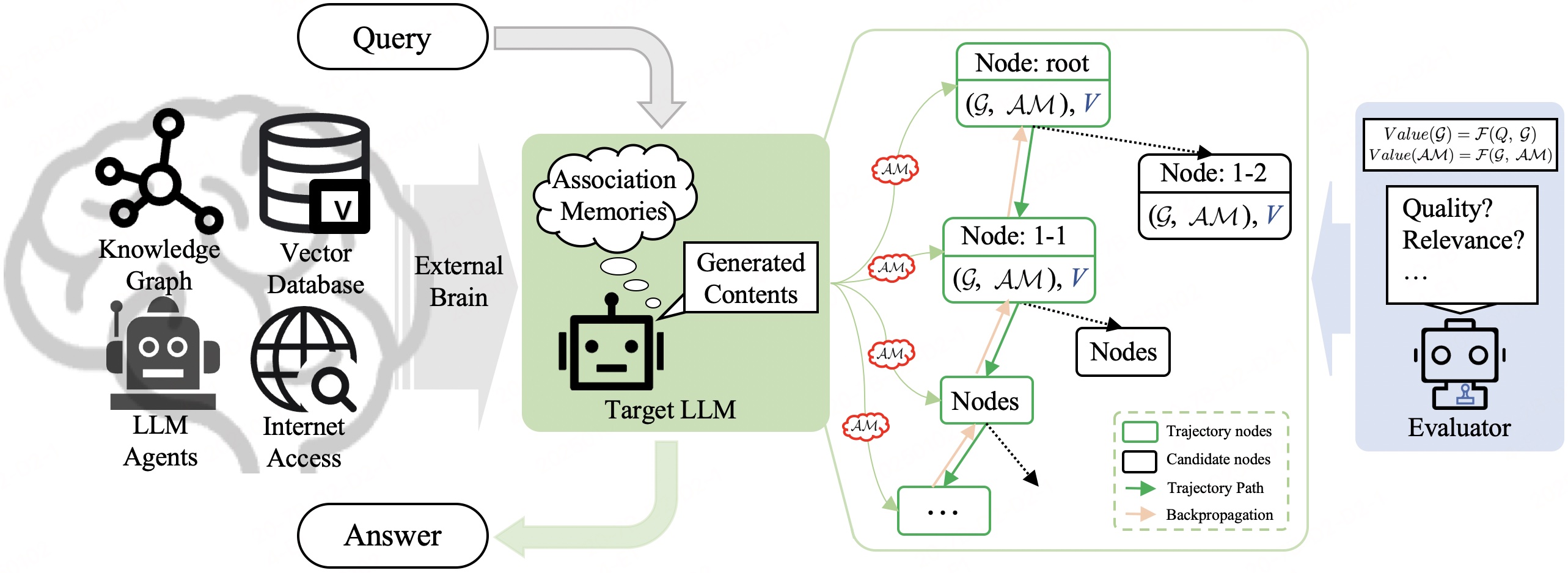}
    \caption{Overview of CoAT framework. The Associative Memory (AM) will be added into each node during reasoning. The ``External Brain ($\mathbb{EB}$)'' is an optional measure to further improve the quality of reasoning results.}
    \label{fig:framework}
\end{figure*}

\section{Related Work}

The development of Large Language Models (LLMs) has witnessed significant advances in recent years, with a particular focus on improving reasoning capabilities. This section reviews key research on LLM inference strategies, the integration of iterative reasoning frameworks, and associative memory mechanisms, all of which inform the design of our Chain-of-Associated-Thoughts (CoAT).

\vspace{3pt}
$\textbf{LLM Inference Strategies}$ Traditional LLMs, including BERT~\cite{devlin2018bert}, GPT-3~\cite{brown2020language} and its successors (like GPT-4~\cite{achiam2023gpt}) rely on a single-shot or few-shot inference paradigm. These methods emphasize the model’s ability to provide accurate responses using fixed prompts, often resulting in outputs that lack robustness in scenarios that require deeper reasoning. To address these limitations, researchers have explored chain-of-thought (CoT) prompting~\cite{wei2022chain} and interleaving retrieval with chain-of-thought (IRCoT)~\cite{trivedi2022interleaving}, which enable LLMs to decompose complex problems into smaller sequential steps. Although this improves reasoning quality, it remains inherently static as the model cannot revisit or refine previous inferences during the reasoning process.

More recently, the variants of CoT, such as self-consistency chain-of-thought (CoT-SC)~\cite{wang2022self} have introduced diversity in reasoning by sampling multiple outputs and selecting the most consistent solution, Graph-of-thought (GoT)~\cite{besta2024graph} has been improved with search algorithms that can search solution paths more effectively, and Tree-of-thought (ToT)~\cite{yao2024tree} prompting uses DFS or BFS search guided by LLMs. However, these methods do not fundamentally alter the underlying inference mechanism, leaving room for further exploration of dynamic and iterative reasoning processes.

The concept of `slow thinking'~\cite{de2024system} has gained traction as an alternative to traditional inference paradigms, inspired by the human ability to deliberate and refine thoughts over time. OpenAI-o1~\cite{jaech2024openai} has been a pioneering framework in this space, demonstrating the benefits of iterative reasoning for tasks involving complex problem solving and decision making. By allowing LLMs to reassess previous steps and integrate new information, slow thinking frameworks improve adaptability and output quality.
These advancements highlight the potential of moving beyond static reasoning toward more dynamic, context-aware methodologies.

\vspace{3pt}
$\textbf{Monte Carlo Tree Search in Inference}$ MCTS has a long history of success in domains requiring decision making under uncertainty, such as game playing~\cite{silver2016mastering} and planning~\cite{coulom2006efficient}. Its ability to balance exploration and exploitation makes it a compelling candidate for enhancing LLM reasoning. Existing works, like LLM-MCTS~\cite{zhao2024large}, LLM agent tree search (LATS)~\cite{zhou2023language} and reasoning via planning (RAP)~\cite{hao2023reasoning}
have integrated MCTS into specific AI systems to improve search space exploration, but its application in LLMs remains limited. Our CoAT extends this approach by leveraging MCTS not only for structured exploration but also as a means to iteratively refine reasoning pathways by inserting associative memory during inference.

\vspace{3pt}
$\textbf{External Knowledge Augmented Mechanisms}$ Augmented knowledge, an external information retrieval process that enables humans to form and retrieve connections between related concepts when thinking, has inspired various machine learning models. Memory-augmented neural networks~\cite{santoro2016meta} and recurrent memory-based architectures~\cite{zaremba2014recurrent} have demonstrated their effectiveness in tasks requiring long-term context retention. However, these systems often lack the flexibility to adapt to evolving information during LLM inference. 

Recent advancements~\cite{gao2023retrieval,yu2023chain,shao2023enhancing,chen2024benchmarking,fan2024survey}, such as native Retrieval Augmented Generation (NativeRAG)~\cite{lewis2020retrieval}, Knowledge Augmented Generation (KAG)~\cite{liang2024kag} and hippocampal indexing RAG (HippoRAG)~\cite{gutierrez2024hipporag}, have addressed this by incorporating external knowledge from vector database or knowledge graph at input stage. CoAT framework builds upon this foundation by introducing a dynamic associative memory mechanism that not only retrieves relevant information but also updates and integrates new knowledge in real time during the reasoning stage without requiring post-training.
Similarly, Search-R1~\cite{jin2025search}, which is conceptually aligned with CoAT, introduces adaptive retrieval capabilities via the reinforcement learning process, which leads to increased computational costs.

Although existing research has made substantial strides in enhancing the reasoning capabilities and adaptability of LLMs, some challenges remain. Static inference strategies and the limited integration of iterative mechanisms continue to constrain the capacity of LLMs to effectively address increasingly complex and dynamic reasoning tasks.
To address these challenges, our proposed CoAT framework synergistically integrates the structured exploration offered by MCTS and the adaptive capabilities of associative memory.

\begin{figure*}[t]
    \centering
    \includegraphics[width=0.85\textwidth]{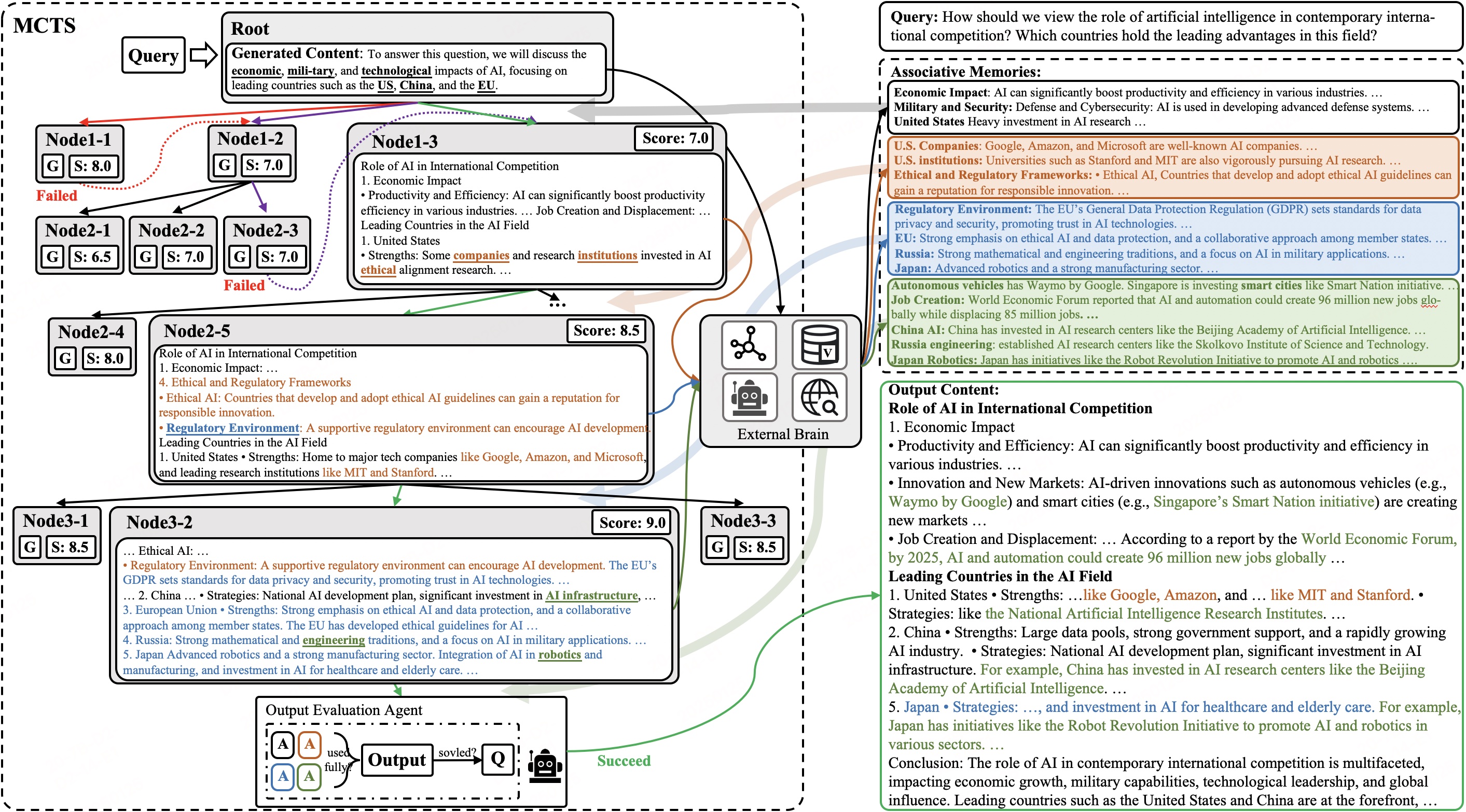}
    \caption{The detailed reasoning process of the CoAT framework. The number of candidate nodes was set to 3.}
    \label{fig:flow-eg}
\end{figure*}

\section{Methodology}
\label{sec:method}

Inspired by the human ability to form associations during cognitive processes and the demonstrated effectiveness of MCTS algorithm in enhancing the reasoning capability of LLMs, we propose the CoAT reasoning framework, as illustrated in Figure~\ref{fig:framework}. The framework leverages the association mechanism to enable LLMs to perform real-time retrieval of relevant information and self-augmentation during the reasoning process. The realization of this functionality is underpinned by our optimized MCTS algorithm, which systematically integrates associative content and generated content through tree node search. By assigning precise values to each node based on our predefined rules, the algorithm facilitates the automatic association process, thereby completing the reasoning task.
To further enhance the reasoning quality of CoAT framework, we have designed a flexible mechanism for sourcing associative content. This mechanism allows the model to either perform self-association or retrieve associative information through external knowledge sources, referred to as an ``External Brain ($\mathbb{EB}$)''. The external brain encompasses commonly used resources such as knowledge graph, vector database, LLM agents, and web search engines. A detailed search process of the CoAT framework when query ``How should we view the role of artificial intelligence in contemporary international competition? Which countries hold the leading advantages in this field?'' is shown in Figure~\ref{fig:flow-eg}.


\subsection{Associative Memory Mechanism}
\label{sec:intro-am}

We introduce associative memory mechanism in the CoAT framework, can be regarded as a novel external knowledge augmentation mechanism, which enables the reasoning process of LLMs to dynamically update and integrate newly retrieved information in real time according to the generated content of each node.
Existing methods primarily focus on incorporating extended knowledge into the reasoning process at its initial stage. However, this approach may lead to incorporation of overly broad knowledge, which introduces two significant drawbacks: (a) an excess of irrelevant information that compromises inference efficiency, and (b) insufficient inclusion of critical content, ultimately degrading inference quality. In contrast, our proposed real-time association mechanism, integrated into the inference process, effectively addresses these issues by dynamically aligning relevant knowledge with the ongoing inference.

The associative memory mechanism generates content that is beneficial for reasoning and has not been previously mentioned in historical contents. The associative content should exhibit minimal redundancy with existing generated contents and should be concise enough to avoid interfering with the reasoning process. Furthermore, the subject of associative content must maintain a strong relevance to the overall reasoning framework. If these conditions are not satisfied, the associative content for the node can be left empty. The above principle will be applied in evaluation stage for evaluating the quality of associative memory.



When generating the associative memory of a node $n_i$, the ``External Brain'' can serve as an alternative approach to enhance the quality of inference results. However, this approach may reduce the efficiency of the inference. This process can be summarized as follows:
\begin{equation}
\label{eq:gen-am}
    \mathcal{AM}(n_i) = \mathbb{EB} \mapsto \mathcal{LLM}(Q~|~\mathcal{G}(n_i)).
\end{equation}
where $\mathcal{G}(n_i)$ denotes the content generated from node $n_i$ and $\mathbb{EB}$ is the External Brain. 

Then, a node can reference both the historical content and the associative memories derived from all of its ancestral nodes. Their historical content and associative content together constitute the comprehensive thinking process of the target LLM. The generation process of each node $n_{i+1}$ is formulated as follows:
\begin{equation}
\label{eq:gen-gen}
    \mathcal{G}(n_{i+1}) = \mathcal{LLM}(Q~|~\mathcal{G}(n_i)~|~\mathcal{AM}(n_{1:i})).
\end{equation}
where $Q$ is the input query and $\mathcal{AM}(n_{1:i})$ denotes the associative memories of nodes $n_1\sim n_i$ in the reasoning trajectory.

\begin{wrapfigure}{l}{0.65\linewidth}
    \centering
    \includegraphics[width=0.9\linewidth]{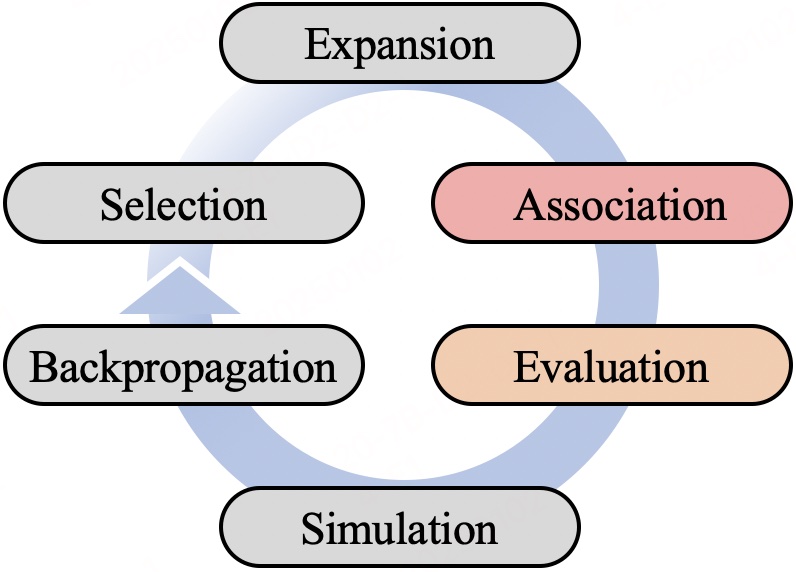}
    \caption{The optimized MCTS process in CoAT.}
    \label{fig:mcts}
\end{wrapfigure}

\subsection{Optimized MCTS}
\label{sec:opt-mcts}

The standard process of the MCTS algorithm consists of four stages: Selection, Expansion, Simulation, and Backpropagation. In the selection stage, MCTS applies the UCT algorithm (Upper Confidence bounds applied to Trees)~\cite{kocsis2006bandit} to choose the best node and then adds it to the trajectory. The UCT of a node $n$ is calculated as follows:
\begin{equation}
\label{eq:uct}
    UCT(n) = V(n) + w\sqrt{\frac{lnN(p)}{N(n)}}.
\end{equation}
where $N(n)$ is the number of visits to node $n$, $V(n)$ is the score value, and $p$ is the parent node of node $n$. $w$ is the exploration weight and is set to 1.0 during CoAT reasoning. When the end of an episode is reached, a back-propagation is carried out to update the value of node $n$ and its parent nodes.

The traditional MCTS algorithm has demonstrated significant success in various decision-making domains. Recently, with advancements in LLM, numerous novel variants of MCTS have been proposed to enable a more effective integration with LLMs. The work of LATS~\cite{zhou2023language} introduces an Evaluation stage after Expansion and a Reflection stage at the end of the process. The evaluation stage assesses the quality of the content generated during the expansion stage, while the reflection stage determines whether the output correctly addresses the inputs. Building on these improvements, we propose an Association stage to simulate the human associative mechanism between the expansion and evaluation stages. The optimized MCTS process is shown in Figure~\ref{fig:mcts}. Consequently, the quality of the associative content is also assessed during the evaluation stage.
The evaluation criteria encompass both the quality of the associative content and its correlation with the content generated during the expansion stage, with the goal of preventing excessive associations and mitigating hallucinations.
Now, the evaluation value of each node $n$ has two components: the generated content value and the associative content value. And the node value is calculated as follows:
\begin{equation}
\label{eq:node-value}
\resizebox{0.89\linewidth}{!}{
    $ V(n) = \mathcal{F}_g(Q,\mathcal{G}(n)) + \beta*\mathcal{F}_a(\mathcal{G}(n),\mathcal{AM}(n)) $.
}
\end{equation}
where $\mathcal{G}(n)$, $\mathcal{AM}(n)$ denotes the generated content and the associative content at node $n$, respectively. $\mathcal{F}$ is the evaluation function for generation and association.
$\beta$ is a weighting coefficient used to control the influence of the associative content, and is set to 0.1 in subsequent experiments.

In the backpropagation stage, we update the visit counts and quality evaluations for every node along the trajectory based on the outcomes of the simulation stage from the leaf node to the root node. The calculation of visit counts is formalized as $C(n_{i+1}) = C(n_i) + 1$. And the quality evaluation value of a parent node $n_p$ will be updated with its children nodes $n^i_c$ as follows: 
\begin{equation}
\label{eq:update-value}
    V(n_p)^* = \frac{V(n_p)*C(n_p) + \sum_i^K{V(n^i_c)}}{C(n_p)+K}.
\end{equation}
where $K$ is the number of candidate nodes of each parent node, $C(n_p)$ is the original visit counts of $n_p$. The updated node value $V(n_p)^*$ is used in the UCT algorithm (Eq.~\ref{eq:uct}) to choose the node of the trajectory in the next selection stage.

To more precisely determine when to terminate the MCTS search process, we applied a specialized Reward Model ($\mathcal{RM}$) to evaluate the content generated at the leaf node of the search trajectory. In certain extreme cases, the search process may enter an ambiguous state, leading to inefficiencies. To mitigate this issue, we introduce a hyper-parameter ($D$) to constrain the maximum depth of the tree search. When the search depth surpasses $D$, the process halts, and the best inference result obtained up to that point is returned. Notably, setting $D=-1$ removes any depth limitation, allowing the search to continue until the optimal result is identified. The flow of the above algorithm can be summarized as Algorithm~\ref{alg:coat} list at Appendix~\ref{sec:appendix-1}.




\begin{table*}[h]
    \centering
    \resizebox{0.83\textwidth}{!}{
        \begin{tabular}{l|c|cc|cc|cc}
            \toprule
            \multirow{2}{*}{\textbf{Framework}} & \multirow{2}{*}{\textbf{Model}} & \multicolumn{2}{c|}{\textbf{HotpotQA}} & \multicolumn{2}{c|}{\textbf{2WikiMultiHopQA}} & \multicolumn{2}{c}{\textbf{MuSiQue}} \\
            \cmidrule(lr){3-4} \cmidrule(lr){5-6} \cmidrule(lr){7-8}
            & & EM & F1 & EM & F1 & EM & F1 \\
            \midrule
            NativeRAG & ChatGPT-3.5 & 43.4 & 57.7 & 33.4 & 43.3 & 15.5 & 26.4 \\
            HippoRAG & ChatGPT-3.5 & 41.8 & 55.0 & 46.6 & 59.2 & 19.2 & 29.8 \\
            \midrule
            IRCoT+NativeRAG & ChatGPT-3.5 & 45.5 & 58.4 & 35.4 & 45.1 & 19.1 & 30.5 \\
            IRCoT+HippoRAG & ChatGPT-3.5 & 45.7 & 59.2 & 47.7 & 62.7 & 21.9 & 33.3 \\
            \midrule
            IRCoT+HippoRAG & DeepSeek-V2 (236B) & 51.0 & 63.7 & 48.0 & 57.1 & 26.2 & 36.5 \\
            KAG &  DeepSeek-V2 (236B) & \underline{62.5} & \textbf{76.2} & \underline{67.8} & \underline{76.2} & \textbf{36.7} & \textbf{48.7} \\ 
            \midrule
            KAG & Qwen2.5-32B-Instruction & 56.6 & 72.1 & 65.9 & 75.5 & 21.3 & 31.4 \\
            \textbf{CoAT(Ours)} & Qwen2.5-32B-Instruction & \textbf{69.6} & \underline{74.2} & \textbf{73.1} & \textbf{78.8} & \underline{34.7} & \underline{39.8} \\
            \bottomrule
        \end{tabular}
    }
    \vspace{-5pt}
    \caption{The end-to-end generation performance of different RAG models on three multi-hop Q\&A datasets. The values in \textbf{bold} and \underline{underline} are the best and second best indicators respectively.}
    \label{tab:quan-rag-gen}
\end{table*}

\section{Experiments}

The implementation of our CoAT framework is built upon the $\bf{LangChain}$~\cite{langchain2025} project.
To evaluate the effectiveness of CoAT framework, we designed two types of validation experiments: (a) assessing the qualitative performance of our CoAT framework in conjunction with LLM, via comparative evaluations against baseline models; (b) quantitatively evaluating the CoAT framework against other state-of-the-art reasoning models on both publicly available open-source datasets and custom-constructed complex reasoning benchmarks.

\begin{figure*}[h]
    \centering
    \includegraphics[width=\textwidth]{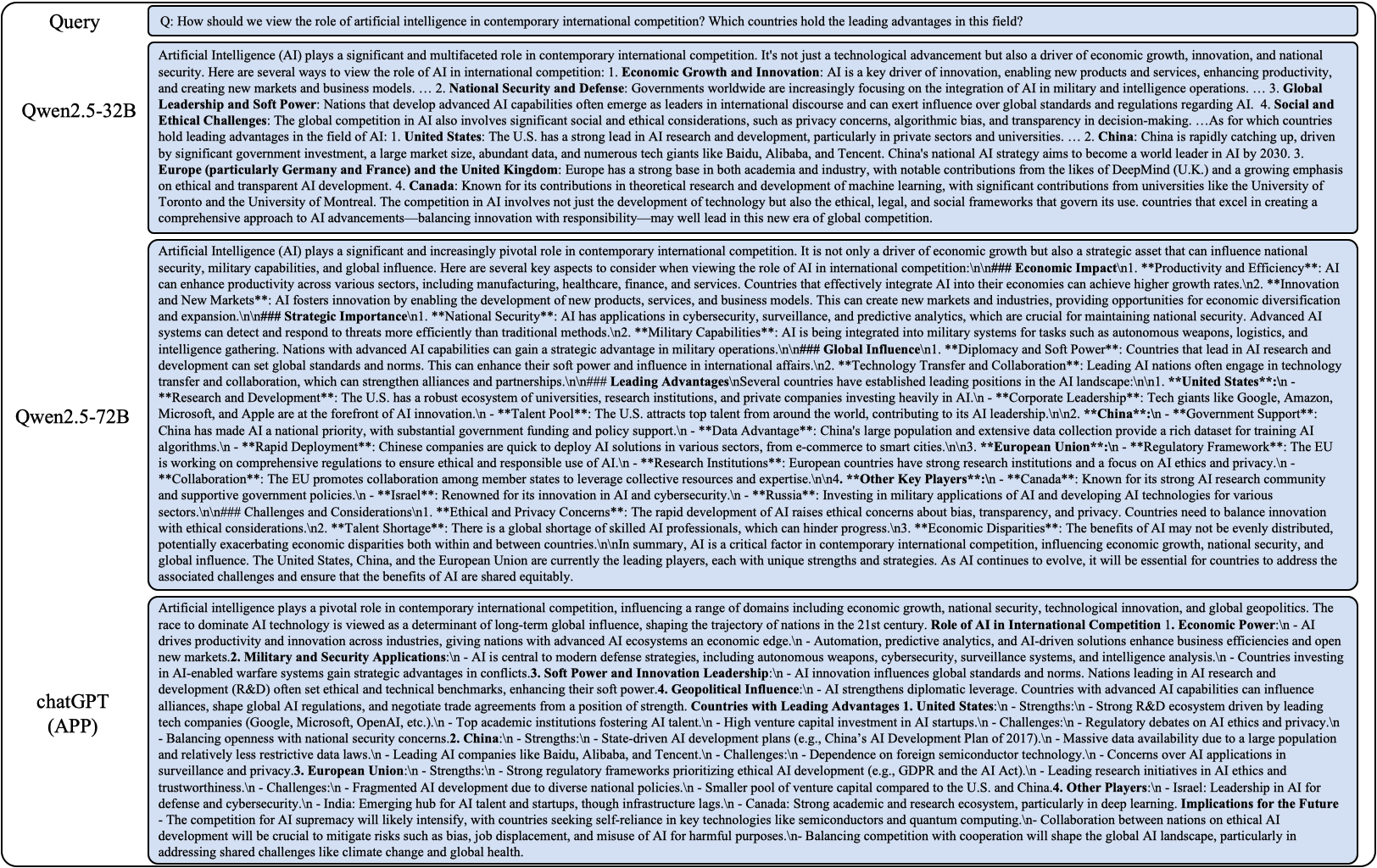}
    \caption{The qualitative results of Qwen2.5-32B-Instruct, ChatGPT (APP) and Qwen2.5-72B-Instruct in a complex reasoning query which need rich associative knowledge.}
    \label{fig:quality-compare}
\end{figure*}

\subsection{Qualitative Performance Evaluation}
\label{sec:quality-exp}



To assess the effectiveness of our CoAT framework in handling real-world reasoning challenges, we designed a series of complex reasoning questions. A case is illustrated in Figure~\ref{fig:quality-compare}. This question requires multidimensional knowledge integration across domains such as economics, ethics.
The CoAT-enhanced model (Qwen2.5-32B) outperforms both the baseline Qwen2.5-32B/72B and ChatGPT models, offering more structured and comprehensive responses. Unlike the baseline outputs, which focus on three to four broad categories, the CoAT model organizes its analysis into five clearly defined dimensions: Economic Impact, Military and Security, Technological Leadership, Ethical and Regulatory Frameworks, and Diplomatic and Soft Power. The additional inclusion of the dimension of ``Ethical and Regulatory Frameworks'' covering AI ethics, privacy regulations, and global governance adds crucial depth and relevance, supported by illustrative examples such as Project Maven and the European AI Alliance.

Moreover, CoAT also demonstrates superior performance in handling the latter part of the task.
While baseline models tend to list countries with limited elaboration, CoAT's output delivers a detailed, evidence-based comparison across six items. Each is analyzed in terms of strengths, strategic priorities, and challenges. For instance, the model highlights India’s AI initiatives for agriculture and urban development, Japan’s robotic-centered AI focus, and Russia’s emphasis on military AI within a constrained geopolitical environment. This granular and policy-relevant analysis underscores CoAT’s strength in supporting nuanced, multi-perspective reasoning, particularly in domains demanding sophisticated geopolitical insight.
The complete response of CoAT can be found in the Appendix~\ref{sec:appendix-2}.

\begin{figure*}[t]
    \centering
    \includegraphics[width=0.8\textwidth]{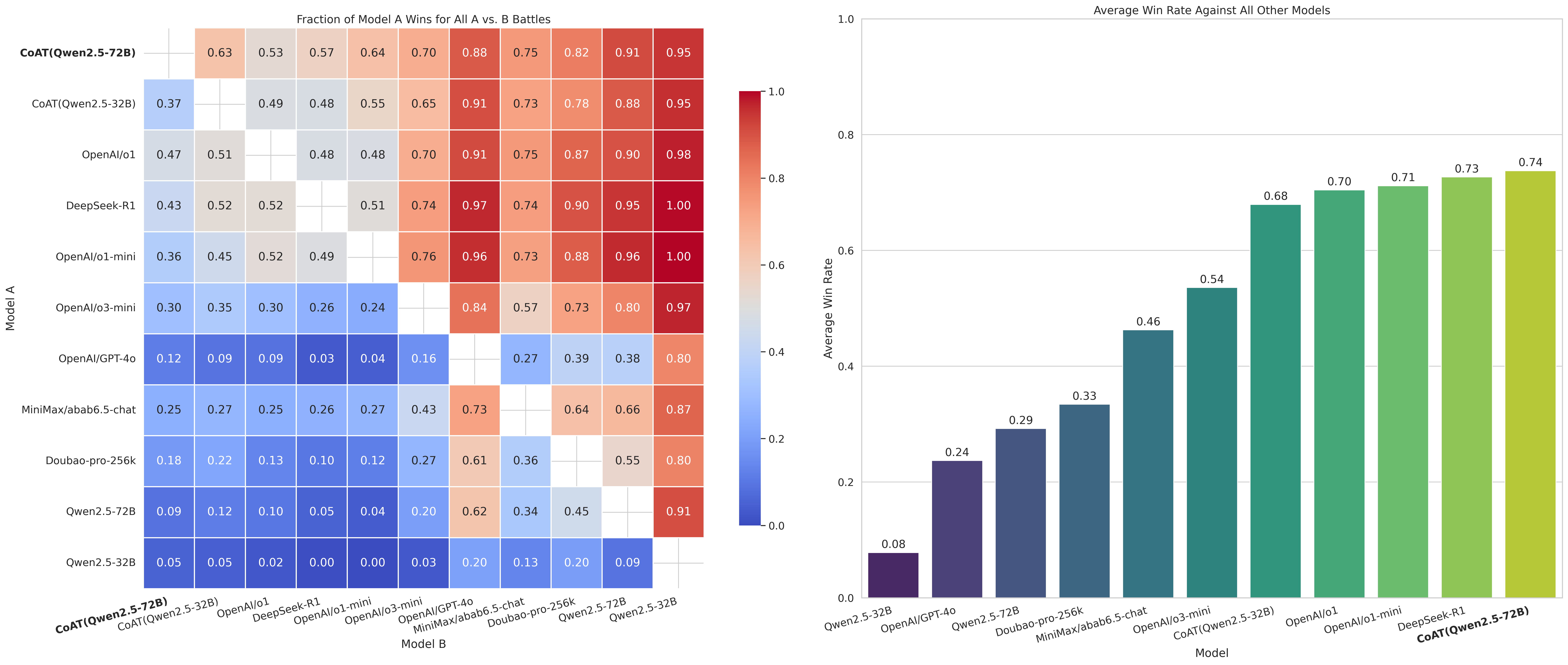}
    \caption{The heatmap of pairwise win rate and the average win rate of all models (Zoom in for best view).}
    \label{fig:win_rate_compare}
\end{figure*}

\subsection{Quantitative Performance Evaluation}

In this section, we will verify the validity of our CoAT framework in two aspects. (a) We compare the base models' reasoning capacity through the CoAT framework with other retrieval-augmented methods on multi-hop datasets. (b) We compare the results of base models through the CoAT framework with other well-known models on a self-built complex comprehensive reasoning dataset.

\textbf{Performance on Multi-hop Datasets}
We enhance the quality of content generated by the associative mechanism through the integration of extended knowledge, and demonstrate that improving the quality of associative content leads to enhanced reasoning ability in our framework. To validate the effectiveness of CoAT framework for the knowledge-intensive question-answering task, we conduct comparative experiments based on retrieval-augmented generation.

The compared methods are NativeRAG~\cite{lewis2020retrieval}, IRCoT~\cite{trivedi2022interleaving}, HippoRAG~\cite{gutierrez2024hipporag}, and KAG~\cite{liang2024kag}. And three widely-used multi-hop QA datasets are HotpotQA~\cite{groeneveld2020simple},
2WikiMultiHopQA~\cite{ho2020constructing},
and MuSiQue~\cite{trivedi2022musique}.

\textbf{Settings.}
For a fair comparison, we follow IRCoT, HippoRAG and KAG utilizing a subset of 1,000 questions from each validation set and constructing a retrieval corpus related to selected questions.
To evaluate QA performance, we adopt two widely used metrics: Exact Match (EM),
and F1 scores.
Furthermore, associative memory is influenced not only by the inherent capabilities of the LLM but also by the quality of retrieval results from external knowledge sources. So we apply the results of KAG's retrieval module as the associative memory during CoAT framework evaluation.


\textbf{Analysis.}
The multi-hop Q\&A performance is presented in Table~\ref{tab:quan-rag-gen}, the results of NativeRAG, HippoRAG and IRCoT using ChatGPT-3.5 and DeepSeek-V2 as the backbone models are excerpts from the official KAG documentation for comparison.
However, since the API service for DeepSeek-V2 has been shut down and its local deployment is also costly, we selected the Qwen2.5-32B-Instruction model as an alternative with comparable capabilities.
Our proposed framework, CoAT, demonstrates significant performance improvements compared to
KAG using the same backbone model, with EM gains of 13.0\%, 7.2\%, and 13.4\% on HotpotQA, 2WikiMultiHopQA, and MuSiQue respectively, and F1 improvements of 2.1\%, 3.3\%, and 8.4\%.
In particular, the performance of our CoAT with Qwen2.5-32B-Instruction is also better than KAG with DeepSeek-V2, with an increase in EM of 7. 1\% and 5. 3\% on HotpotQA, 2WikiMultiHopQA datasets.

The observed performance improvements can be largely attributed to the more comprehensive exploration of semantically related entities during the reasoning process in our framework.
Leveraging the previous retrieved passages, we employ the association mechanism to identify and expand upon salient entities that are essential for multi-hop reasoning.
The retrieval results are further refined in the subsequent content generation stage to enhance response accuracy.
However, the association mechanism will lead the model to generate explanatory contents when a direct answer is unavailable, which can reduce response precision and consequently lower the overall F1 score.

\textbf{Performance on CRB Dataset}
\label{sec:intro-crb}

To better demonstrate the effectiveness of CoAT framework in associative reasoning tasks, we constructed a high-quality reasoning dataset, referred to as the Comprehensive Reasoning Benchmark (CRB). This dataset encompasses various disciplines, including politics, scientific and technological domains, international relations, economics, law, and history, among others. The tasks in this dataset require advanced analytical skills, case-based evidence, and rigorous logical reasoning.

\textbf{Dataset.}
The CRB dataset contains 205 professionally reviewed questions, each accompanied by its corresponding evaluation rules and total score, which together constitute the final evaluation entries. More details are provided in Appendix~\ref{sec:appendix-3} and this dataset has been released in \url{https://github.com/IceSwordLab/CoAT}.

\textbf{Settings.}
Based on the CRB dataset, we designed two series of experiments. First, we selected multiple state-of-the-art generative models (MiniMax/abab6.5-chat, Doubao-pro-256k, OpenAI/GPT-4o, Qwen2.5-32B-Instruct, Qwen2.5-72B-Instruct), reasoning models (DeepSeek-R1, OpenAI/o1, OpenAI/o1-mini, OpenAI/o3-mini) and our CoAT framework with two base models (Qwen2.5-32B-Instruct, Qwen2.5-72B-Instruct) to generate answers for the questions in the dataset. (For a more detailed discussion on model selection, please refer to the Appendix~\ref{sec:appendix-sup1}.) The answers are then evaluated according to Judge Rules, and scores were assigned accordingly. The final average score for each model was computed using a standardized formula:
$S_{\mathcal{M}} = \frac{1}{N}\sum_i^N{(\frac{s_i}{s_T})}$,
where $s_i$ and $s_T$ are the evaluated score and the maximum score, respectively.
The experimental results are presented in Table~\ref{tab:quan-crb}.

Second, we conducted pairwise comparisons of the responses generated by the above models to assess which model's responses demonstrate greater comprehensiveness and depth of detail. Based on these comparisons, we derived the win rate heatmap and the average win rate for each model. The results of this experiment are illustrated in Figure~\ref{fig:win_rate_compare}.

\begin{table}[h]
    \centering
    \resizebox{0.75\linewidth}{!}{
        \begin{tabular}{cc}
            \toprule
            Models & Average Scores \\
            \midrule
            Qwen2.5-32B-Instruct & 0.55 \\
            OpenAI/GPT-4o & 0.59 \\
            Doubao-pro-256k & 0.61 \\
            Qwen2.5-72B-Instruct & 0.62 \\
            MiniMax/abab6.5-chat & 0.66 \\
            \midrule
            OpenAI/o3-mini & 0.64 \\
            OpenAI/o1-mini & 0.71 \\
            OpenAI/o1 & 0.73 \\
            DeepSeek-R1 & 0.75 \\
            \midrule
            \textbf{CoAT (Qwen2.5-32B)} & \underline{0.77} \\
            \textbf{CoAT (Qwen2.5-72B)} & \textbf{0.80} \\
            \bottomrule
        \end{tabular}
    }
    \vspace{-5pt}
    \caption{The results of all comparison models on CRB.}
    \label{tab:quan-crb}
\end{table}

\textbf{Analysis.}
The results presented in Table~\ref{tab:quan-crb} support the following conclusions. First, our CoAT framework significantly improves the performance of the base models, Qwen2.5-32B-Instruction and Qwen2.5-72B-Instruction, with relative gains of 22\% and 18\% in the evaluated scores, respectively.
This performance gain can be attributed to two key components: the entity association enabled by the association mechanism, and the progressive exploration strategy of MCTS, which together enhance both the comprehensiveness and accuracy of the generated answers.
Second, by leveraging APIs of smaller-scale language models within our CoAT framework, enhanced reasoning performance can be achieved without the need for additional model training or fine-tuning.
Finally, reasoning models tend to outperform generative models in CRB datasets, highlighting the advantages of structured reasoning in complex benchmarks.


As illustrated in Figure~\ref{fig:win_rate_compare}, the Qwen2.5-72B-Instruct model integrated with our CoAT reasoning framework outperforms other models in pairwise evaluations. Specifically, the Qwen series achieves a 50\% relative improvement in the average win rate when using CoAT, indicating the effectiveness of our proposed framework.
Moreover, generative models augmented with the CoAT framework can achieve performance comparable to that of dedicated reasoning models.

\subsection{Ablation Experiment}
\label{sec:abla-exp}


\textbf{Effectiveness of $\mathcal{AM}$.}
To separately verify the effectiveness of the associative memory mechanism (Sec.~\ref{sec:intro-am}) and the optimized value computation for MCTS nodes (Sec.~\ref{sec:opt-mcts}), we performed three experiments using the CRB dataset, and calculated the resulting scores and win rates for both settings:
1) We applied the CoAT framework to generate results either with $\mathcal{AM}$ integrated into each node or without it;
2) We incorporated only the content of $\mathcal{AM}$ and without considering its quality.
3) We conducted a series of experiments with different values of $\beta$ in Eq.~\ref{eq:node-value} to explore the optimal solution.
In Appendix~\ref{sec:appendix-sup2}, we provide an additional detailed discussion of the countermeasures for $\mathcal{AM}$ under extreme situations, aiming to enhance the efficiency of the association process.

\begin{table}[h]
    \centering
    \resizebox{0.85\linewidth}{!}{
        \begin{tabular}{ccc}
            \toprule
            Models & Versions & Scores \\
            \midrule
            \multirow{3}{*}{CoAT (Qwen2.5-32B)} & w/o $\mathcal{AM}$ & 0.67 \\
            & w/ $\mathcal{AM}$ \& $\beta$=0 & 0.75 \\
            & w/ $\mathcal{AM}$ \& $\beta$=0.1 & \underline{0.77} \\
            \midrule
            \multirow{3}{*}{CoAT (Qwen2.5-72B)} & w/o $\mathcal{AM}$ & 0.71 \\
             & w/ $\mathcal{AM}$ \& $\beta$=0 & \underline{0.77} \\
             & w/ $\mathcal{AM}$ \& $\beta$=0.1 & \textbf{0.80} \\
            \bottomrule
        \end{tabular}
    }
    \vspace{-5pt}
    \caption{The results of the above settings on CRB.}
    \label{tab:quan-setting}
\end{table}

\begin{figure}
    \centering
    \includegraphics[width=0.85\linewidth]{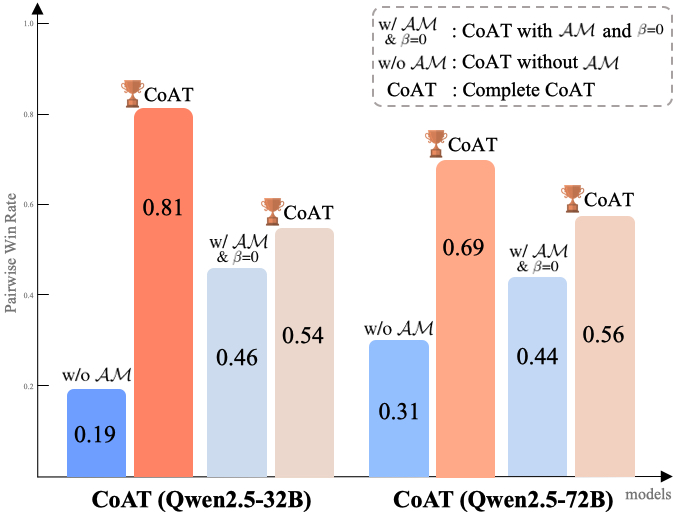}
    \vspace{-10pt}
    \caption{The pairwise win rates.} 
    \label{fig:wo-am}
\end{figure}

\begin{figure}[t]
    \centering
    \includegraphics[width=0.85\linewidth]{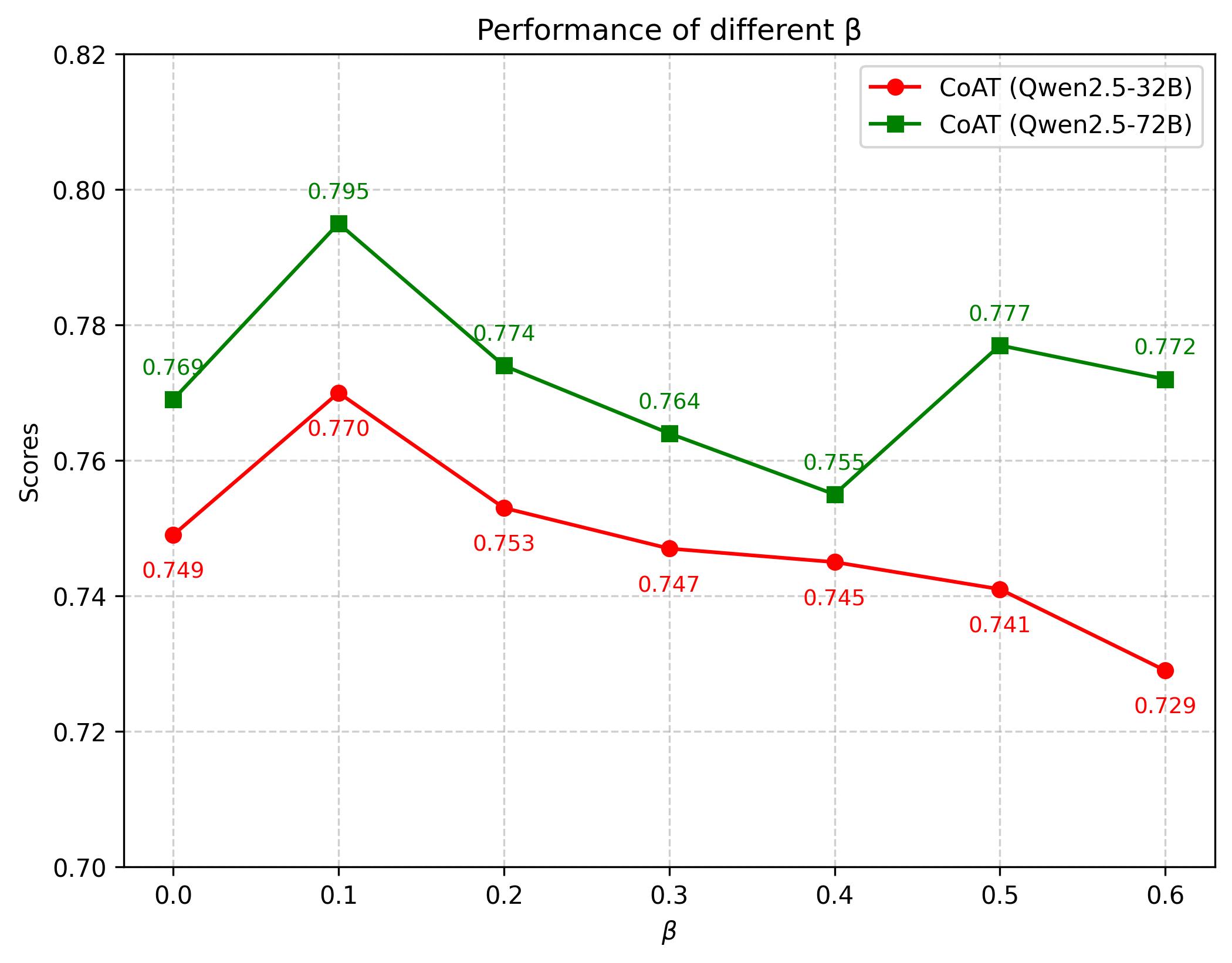}
    \vspace{-10pt}
    \caption{The results of different values of $\beta$.} 
    \label{fig:beta-values}
\end{figure}

\textbf{Settings.}
1) \textbf{w/o $\mathcal{AM}$:} We disabled the generation of $\mathcal{AM}$ at each node as defined in Eq.~\ref{eq:gen-am} and set its value to empty in Eq.~\ref{eq:gen-gen}. Subsequently, we omitted the contribution of $\mathcal{AM}$ in Eq.~\ref{eq:node-value}.
2) \textbf{w/ $\mathcal{AM}$\&$\beta$=0:} In Eq.~\ref{eq:node-value}, $\beta$ is set to 0.0, while all other components are identical to those in the complete CoAT.
The visual results of the above settings are shown in Table~\ref{tab:quan-setting} and Figure~\ref{fig:wo-am}. The results of pairwise comparisons between the above settings and all base models are provided in Appendix~\ref{sec:appendix-4}.
3) The values of $\beta$ are sampled from [0, 0.6], and the results are shown in Figure~\ref{fig:beta-values}.

\textbf{Analysis.}
The results in Table~\ref{tab:quan-setting} show that incorporating the whole $\mathcal{AM}$ mechanism led to an approximately 10\% improvement in average score compared to the baseline without $\mathcal{AM}$. Moreover, the comparison results in Figure~\ref{fig:wo-am} further validate the performance improvements achieved through the integration of $\mathcal{AM}$.
Figure~\ref{fig:beta-values} shows that although the value of $\beta$ has little impact on the final score, the optimal result is achieved within the entire range when $\beta=0.1$.

\section{Conclusion}

In this paper, we proposed the CoAT reasoning framework, which advances LLM reasoning by integrating an optimized MCTS algorithm and a dynamic associative memory mechanism. These innovations enable structured exploration of reasoning pathways and adaptive knowledge updating, addressing limitations of generative LLMs. The experimental results demonstrated that CoAT outperforms other models in accuracy, coherence, and diversity. Our work highlights the potential of combining structured search and adaptive associative memory in LLMs, offering a new exploration for future research on integrating external real-time knowledge for real-world applications.


\section*{Limitations}
Although our framework outperforms baseline models, there is still room for improvements. Due to the expansion of the search space and the incorporation of novel associative memory, our framework achieves more comprehensive content generation than the baseline models. However, such improvement comes at the cost of increased reasoning time.
Additionally, despite the considerable manual efforts invested in curating the self-constructed CRB dataset, there is still room for further quality refinement.


\section*{Ethics Statement}
Our research focuses on enhancing the reasoning capacity of LLMs. There are no specific ethical concerns directly associated with this work.
However, we recognize and emphasize the ethical mindfulness throughout our research. In particular, during the construction of the CRB dataset, no ethical guidelines were violated, and careful attention was paid to data quality and integrity. The broader impact of our work lies in advancing the performance of baseline models, thereby contributing to the improvement of the quality and reliability of content generated by LLMs.
All the datasets and models used in this work are publicly available with permissible licenses.

\bibliography{custom}

\clearpage
\appendix
\onecolumn
\section{Appendix}
\label{sec:appendix}

\subsection{CoAT Reasoning Algorithm}
\label{sec:appendix-1}
The flow of our CoAT reasoning algorithm is summarized in Algorithm~\ref{alg:coat}. The details of the CoAT algorithm are described in Section~\ref{sec:opt-mcts}.

\begin{algorithm*}[h]
    \caption{CoAT Reasoning Algorithm}
    \label{alg:coat}
    \textbf{Require}: Input query ($Q$), Evaluate function ($\mathcal{F}$), Reward Model ($\mathcal{RM}$), Number of candidate nodes ($K$), Search Depth ($D$), Exploration weight ($w$), $\mathcal{AM}$ impact factor ($\beta$)
    \begin{algorithmic}[1] 
        \STATE Initialize: Root node $n\leftarrow{(\mathcal{G}, \mathcal{AM})}$, $d\leftarrow{1}$ 
        \STATE Judge: $\mathcal{RM}(\mathcal{G})$ is completed for $Q$.
        \WHILE{Uncompleted \textbf{and} $d<D$}
        \STATE Selection: choose the best trajectory by UCT.
        \FOR{Child Node $i\leftarrow{1,...,K}$}
        \STATE Generation ($\mathcal{G^*}$): $n_i\leftarrow{\mathcal{LLM}(Q|\mathcal{G}|\mathcal{AM})}$.
        \STATE Association ($\mathcal{AM^*}$): $n_i\leftarrow{\mathcal{LLM}(Q|\mathcal{G^*})}$.
        \STATE Evaluation ($V$): $n_i\leftarrow{\mathcal{F}(Q,\mathcal{G^*}) + \beta\mathcal{F}(\mathcal{G^*},\mathcal{AM^*})}$.
        \ENDFOR
        \STATE Judge: $\mathcal{RM}(\mathcal{G^*})$ of the best child is completed for $Q$.
        \IF {Completed}
        \STATE \textbf{return} $\mathcal{G^*}$.
        \ELSE
        \STATE Backpropagation: value of $n_p\leftarrow{Equation~\ref{eq:update-value}}$.
        \STATE Update: visit ($V^*$) of $n_p\leftarrow{C(n_p)+K}$.
        \STATE Update: $d\leftarrow{d + 1}$.
        \ENDIF
        \ENDWHILE
        \STATE \textbf{return} $\hat{\mathcal{G}}$ of the best trajectory.
    \end{algorithmic}
\end{algorithm*}

\FloatBarrier 

\subsection{Selection of Comparative Models}
\label{sec:appendix-sup1}
During the CoAT verification stage, extensive experiments were required, which resulted in a large volume of model access. Considering cost-effectiveness, we prioritized using open-source models. For our task, our primary focus was on the generation and summarization capabilities of LLMs. We researched several mainstream open-source models available at the time. Both the Qwen2.5 series and the Llama3.1 series exhibited comparable performance in these aspects. The detailed comparison results are available on the official Qwen2.5 Blog\footnote{\url{https://qwenlm.github.io/blog/qwen2.5}}. However, the Qwen series offers broader multilingual support, which is particularly beneficial for the associative memory module, as it may require the understanding of materials in various languages. According to their official Hugging Face model cards, Qwen2.5 supports more than 29 languages\footnote{\url{https://huggingface.co/Qwen/Qwen2.5-72B}}, whereas Llama 3.1 supports 8 languages\footnote{\url{https://huggingface.co/meta-llama/Llama-3.1-70B}}. Therefore, the Qwen series was deemed more appropriate for our experiments. However, other LLMs (such as the Llama series) can be seamlessly integrated into the CoAT framework simply by replacing the service API or endpoint.

\clearpage
\subsection{Qualitative Performance Exhibition}
\label{sec:appendix-2}


The input of all LLMs and CoAT are only the query questions with no extra instruction prompts and system prompts.
The responses of three comparison models for a query (How should we view the role of artificial intelligence in contemporary international competition? Which countries hold the leading advantages in this field?) are shown in Figure~\ref{fig:quality-compare}. And the corresponding response of CoAT (Qwen2.5-32B-Instruct) is shown as follows:
\begin{lstlisting}
Role of AI in International Competition
 1. **Economic Impact**
    - **Productivity and Efficiency**: AI can significantly boost productivity and efficiency in various industries. For instance, in manufacturing, AI can optimize supply chains and reduce waste. In healthcare, AI-powered diagnostic tools can improve patient outcomes and reduce costs.
    - **Innovation and New Markets**: AI-driven innovations such as autonomous vehicles (e.g., Waymo by Google) and smart cities (e.g., Singapore's Smart Nation initiative) are creating new markets and business opportunities.
    - **Job Creation and Displacement**: While AI can create new jobs in technology and data science, it can also displace traditional jobs. According to a report by the World Economic Forum, by 2025, AI and automation could create 96 million new jobs globally while displacing 85 million jobs.
    
 2. **Military and Security**
    - **Defense and Cybersecurity**: AI is used in developing advanced defense systems. For example, the U.S. Department of Defense has projects like Project Maven, which uses AI for image recognition in drone footage. China has also invested in AI for military applications, including autonomous drones and cyber defense systems.
    - **Intelligence and Surveillance**: AI can improve intelligence gathering and surveillance. The U.S. uses AI for analyzing satellite imagery and detecting potential threats. China has deployed AI in surveillance systems, such as facial recognition technology in public spaces.
    - **Strategic Decision-Making**: AI can assist in strategic decision-making. For example, the U.S. military uses AI to analyze data and provide predictive insights for military operations.
    
 3. **Technological Leadership**
    - **Research and Development**: Leading countries invest heavily in AI R&D. For example, the U.S. has a robust AI research ecosystem with significant funding from both the government and private sector. China has a national AI development plan with substantial government support.
    - **Patent Filings and Publications**: Leading countries often have a high number of AI-related patent filings and academic publications. According to the World Intellectual Property Organization (WIPO), China and the U.S. are the top two countries in AI patent filings.
    - **Talent Attraction and Retention**: Attracting and retaining top AI talent is crucial. Countries like the U.S. and Canada offer attractive research environments and job opportunities. For example, Canada's Vector Institute is a leading AI research center that attracts top talent from around the world.
    
 4. **Ethical and Regulatory Frameworks**
    - **Ethical AI**: Countries that develop and adopt ethical AI guidelines can gain a reputation for responsible innovation. For example, the EU has developed ethical guidelines for AI, emphasizing issues like transparency, fairness, and accountability.
    - **Regulatory Environment**: A supportive regulatory environment can encourage AI development. The EU's General Data Protection Regulation (GDPR) sets standards for data privacy and security, promoting trust in AI technologies.
 
 5. **Diplomatic and Soft Power**
    - **Global Standards and Norms**: Leading countries can influence the development of global standards and norms for AI. The EU, for instance, is working on shaping international AI regulations through initiatives like the European AI Alliance.
    - **AI Diplomacy**: Countries can use AI to enhance their diplomatic efforts. For example, Japan is using AI to improve communication and data-driven decision-making in international negotiations.
    
Leading Countries in the AI Field
 **1. United States**
    - **Strengths**: Home to major tech companies like Google, Amazon, and Microsoft, and leading research institutions like MIT and Stanford.
    - **Strategies**: Heavy investment in AI R&D, establishment of AI labs, and strong private-public partnerships. For example, the U.S. has established AI institutes like the National Artificial Intelligence Research Institutes.
    - **Challenges**: Balancing AI innovation with ethical considerations and ensuring a skilled workforce. The U.S. faces challenges in maintaining a competitive edge in AI talent and addressing ethical concerns.
  
 **2. China**
    - **Strengths**: Large data pools, strong government support, and a rapidly growing AI industry. China has a national AI development plan with significant government investment.
    - **Strategies**: National AI development plan, significant investment in AI infrastructure, and fostering a large AI talent pool. For example, China has invested in AI research centers like the Beijing Academy of Artificial Intelligence.
    - **Challenges**: Ensuring data privacy and ethical use of AI, and addressing international concerns about surveillance and human rights. China faces scrutiny over its use of AI for social control and surveillance.
    
 **3. European Union**
    - **Strengths**: Strong emphasis on ethical AI and data protection, and a collaborative approach among member states. The EU has developed ethical guidelines for AI and promotes data privacy through regulations like GDPR.
    - **Strategies**: Development of the General Data Protection Regulation (GDPR) and the European AI Alliance. For example, the EU has launched the European AI Alliance to promote ethical AI development.
    - **Challenges**: Harmonizing AI policies across member states and competing with the US and China in terms of investment and innovation. The EU faces challenges in coordinating AI policies among its diverse member states.
  
 **4. Russia**
    - **Strengths**: Strong mathematical and engineering traditions, and a focus on AI in military applications. Russia has a strong research base in mathematics and computer science.
    - **Strategies**: Government investment in AI for defense and security, and development of AI in critical sectors like energy and space. For example, Russia has established AI research centers like the Skolkovo Institute of Science and Technology.
    - **Challenges**: Limited private sector involvement and international sanctions that can restrict access to advanced technologies. Russia faces challenges in fostering a vibrant AI ecosystem and accessing cutting-edge technologies.

 **5. Japan**
    - **Strengths**: Advanced robotics and a strong manufacturing sector. Japan is a leader in robotics and has a strong manufacturing base.
    - **Strategies**: Integration of AI in robotics and manufacturing, and investment in AI for healthcare and elderly care. For example, Japan has initiatives like the Robot Revolution Initiative to promote AI and robotics in various sectors.
    - **Challenges**: Aging population and the need for more AI talent. Japan faces challenges in addressing the needs of its aging population and attracting a sufficient number of AI professionals.
  
 **6. India**
    - **Strengths**: Large pool of skilled IT professionals and a growing tech ecosystem. India has a strong IT industry and a large number of skilled professionals.
    - **Strategies**: National AI strategy focusing on healthcare, agriculture, education, and smart cities. For example, India has launched initiatives like the National Artificial Intelligence Program to promote AI development.
    - **Challenges**: Infrastructure gaps and the need for significant investment in AI research and development. India faces challenges in building a robust AI infrastructure and attracting investment.
    
 
Conclusion
    The role of AI in contemporary international competition is multifaceted, impacting economic growth, military capabilities, technological leadership, and global influence. Leading countries such as the United States and China are at the forefront, with significant investments in AI R&D, strong private-public partnerships, and robust talent ecosystems. Other nations like the EU, Russia, India, and Japan are also making significant strides in the AI field, each with their unique strengths and challenges. The strategic positioning of these countries in the AI landscape is crucial for shaping the future of global competition and innovation.
\end{lstlisting}
The detailed introduction and analysis of the results are described in Section~\ref{sec:quality-exp}.

\FloatBarrier

\subsection{Introduction of CRB Dataset}
\label{sec:appendix-3}
The data entry in the CRB\footnote{\url{https://github.com/IceSwordLab/CoAT}} is structured referring to the design principles of subjective questions in the Chinese Gaokao examination. Specifically, each entry consists of three components: Question, Judge Rules, and Score. Judge Rules outline a series of fundamental key points that must be addressed to provide an adequate response.   Each key point corresponds to a specific score, and the inclusion of these key points in an answer results in the allocation of the corresponding score. Additionally, Judge Rules incorporate higher-level criteria as bonus points. Score assigned to each data entry represents the maximum attainable score for that entry. Ultimately, we selected 205 professionally reviewed entries as the final test dataset.
An example entry from the CRB dataset is shown below: 

\begin{lstlisting}
{
    "question": "What are the key challenges in achieving global carbon neutrality by 2050? Which countries are making the most progress?",
    "judge_rules": "Evaluation Criteria:\n1. (This section is worth 10 points) Discuss the main challenges of carbon neutrality, including technological bottlenecks, economic burden, policy coordination, energy transition, and social acceptance. One point is awarded for addressing each aspect. An additional point can be earned for each aspect if it includes at least one real-world example.\n2. (This section is worth 10 points) Discuss global leading countries in carbon neutrality efforts, including the European Union, China, the United States, Japan, and India. One point is awarded for each country discussed.\n   a. An additional point can be earned for each country if the discussion covers advantages, disadvantages, and strategies.\n   b. Another point can be earned if the discussion of each country includes at least one real-world example.\n3. (This section is worth 3 points) One point will be awarded for each of the following: fluent language, detailed discussion, and factual accuracy.",
    "total_score": 23
}
\end{lstlisting}

\begin{figure}[h]
    \centering
    \includegraphics[width=\textwidth]{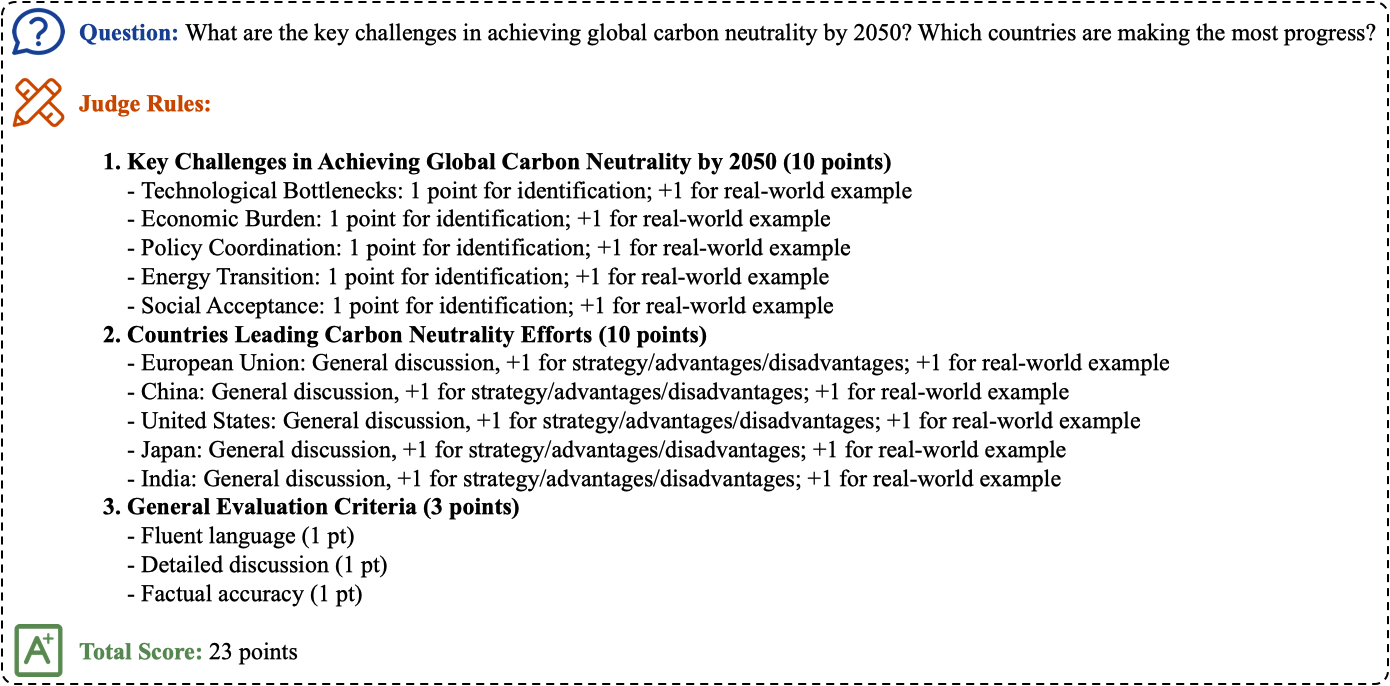}
    \caption{This figure presents the key scoring points identified in the sample question from the Comprehensive Reasoning Benchmark (CRB) dataset.}
    \label{fig:crb-sample}
\end{figure}

Each entry in the CRB dataset consists of three components: Question, Judge Rules and Total Score. The language of Question and Judge Rules is either English or Chinese. The key scoring points of the sample shown above are illustrated in Figure~\ref{fig:crb-sample}. For experiments with the CRB dataset, please refer to Section~\ref{sec:intro-crb}.

In Figure~\ref{fig:crb-sample}, we provide a detailed breakdown of all the scoring points specified in the judge rules for the question. Each judge rule in the CRB dataset has undergone a professional manual review to assess the reasonableness of the scoring points and the accuracy of the total score. However, there may still be room for further improvement.

\FloatBarrier

\subsection{Efficiency Analysis of $\mathcal{AM}$}
\label{sec:appendix-sup2}
In our CoAT framework, the association mechanism conducts necessary associations of key entity information based on the generated content of each node to dynamically improve the quality of the final results. Therefore, associating the memory from the initial stage will only involve important information for the key entity information and will not be particularly cumbersome. However, if there is extremely long content in the form of a long document, we will also perform truncation or segmentation by extracting key information, document summaries, or document chunks to ensure that the associative memory actually used for CoAT inference does not become excessively large. Meanwhile, long documents can be transformed into knowledge graphs using document-to-knowledge graph techniques, such as Docs2KG~\cite{sun2024docs2kg}. Knowledge graphs are inherently supported in CoAT. Moreover, handling long documents is primarily an engineering implementation. The strategies discussed above are capable of addressing this challenge. Therefore, the presence of long texts does not hinder the use of MCTS in CoAT for enhanced reasoning.

Additionally, the following strategies are applicable: (1) Implementing a hierarchical retrieval mechanism by splitting documents into smaller chunks, which allows the EB to return only the most relevant portions for associative reasoning—an approach widely adopted in frameworks like LlamaIndex; and (2) Applying lightweight summarization models such as DistilBART to compress long documents before retrieval, as exemplified in OpenAI’s official guide on ``Retrieval Augmented Generation (RAG) and Semantic Search for GPTs.\footnote{\url{https://help.openai.com/en/articles/8868588-retrieval-augmented-generation-rag-and-semantic-search-for-gpts}}''

\subsection{Comparison Results of CoAT with Different Settings}
\label{sec:appendix-4}
\begin{figure*}[t]
    \centering
    \includegraphics[width=\textwidth]{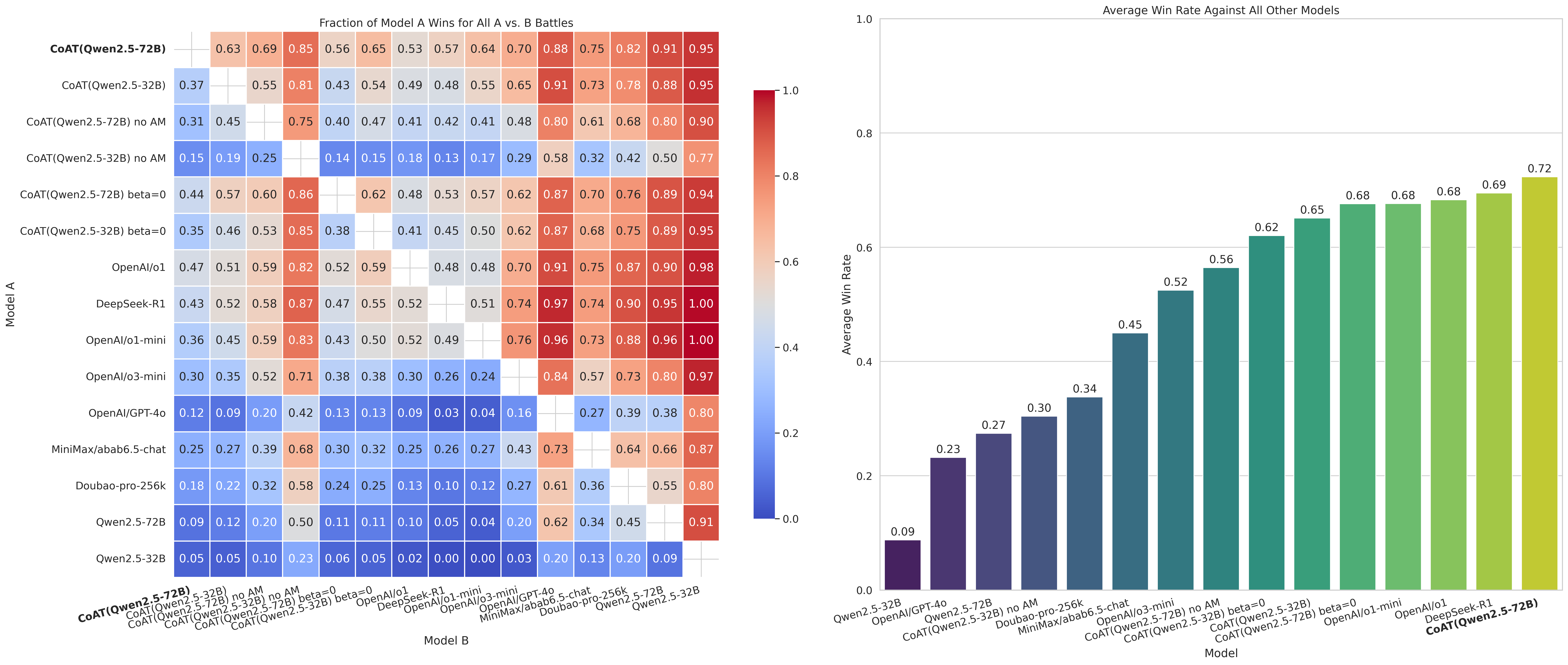}
    \caption{This figure presents the pairwise comparison results among all models. (Zoom in for best view)}
    \label{fig:win_rate_compare-wo-am}
\end{figure*}

The results of the pairwise comparison of the CoAT framework with different settings among all models are shown in Figure~\ref{fig:win_rate_compare-wo-am}. The description of this experiment refers to Section~\ref{sec:abla-exp}.

\FloatBarrier

\end{document}